%% file: main.tex
\documentclass[lettersize,journal]{IEEEtran}
\usepackage{amsmath,amsfonts}
\usepackage{algorithmic}
\usepackage{algorithm}
\usepackage{array}
\usepackage[caption=false,font=normalsize,labelfont=sf,textfont=sf]{subfig}
\usepackage{textcomp}
\usepackage{stfloats}
\usepackage{url}
\usepackage{verbatim}
\usepackage{graphicx}
\usepackage{cite}
\usepackage{graphicx}
\usepackage{amsfonts}
\usepackage{xcolor}
\usepackage{blindtext}
\usepackage{hyperref}
\usepackage{multirow}
\usepackage[switch]{lineno}

\begin{document}

\title{Contrastive Brain Network Learning via Hierarchical Signed Graph Pooling Model}

\author{Haoteng Tang, ~\IEEEmembership{Student Member,~IEEE,} 
        Guixiang Ma, ~\IEEEmembership{Member,~IEEE,}
        Lei Guo, ~\IEEEmembership{Student Member,~IEEE,} \\
        Xiyao Fu, ~\IEEEmembership{Student Member,~IEEE,}
        Heng Huang, ~\IEEEmembership{Member,~IEEE,}
        Liang Zhan$^{\dag}$, ~\IEEEmembership{Member,~IEEE}
\thanks{H. Tang, L. Guo, X. Fu, H. Huang and L. Zhan are with the Department of Electrical and Computer Engineering, University of Pittsburgh, Pittsburgh, PA, 15260, USA (e-mail: \{haoteng.tang, Lei.guo, xy\_fu, heng.huang, liang.zhan\}@pitt.edu)}
\thanks{G. Ma is with the Department of Computer Science, University of Illinois at Chicago, Chicago, IL 60607, USA (e-mail:guixiang.ma@intel.com). This work was done before Dr. Ma joined Intel.}
\thanks{Liang Zhan is the corresponding author (denoted by $\dag$)}
}

\markboth{Journal of Transactions on Neural Networks and Learning Systems,~Vol.~XX, No.~X, XXX~XXXX}%
{Shell \MakeLowercase{\textit{et al.}}: A Sample Article Using IEEEtran.cls for IEEE Journals}


\maketitle

\begin{abstract}
\input{00_abstract}
\end{abstract}


\input{01_introduction}

\input{02_relatedwork}

\input{03_prelim}
\input{04_method}

\input{05_experiment}

\input{06_conclusion}

\IEEEpeerreviewmaketitle
\bibliographystyle{IEEEtran}
\bibliography{reference}

\end{document}

%% file: 00_abstract.tex
Recently brain networks have been widely adopted to study brain dynamics, brain development and brain diseases. 
Graph representation learning techniques on brain functional networks can facilitate the discovery of novel biomarkers for clinical phenotypes and neurodegenerative diseases.
However, current graph learning techniques have several issues on brain network mining. 
Firstly, most current graph learning models are designed for unsigned graph, which hinders the analysis of many signed network data (e.g., brain functional networks). 
Meanwhile, the insufficiency of brain network data limits the model performance on clinical phenotypes predictions. 
Moreover, few of current graph learning model is interpretable, which may not be capable to provide biological insights for model outcomes. 
Here, we propose an interpretable hierarchical signed graph representation learning model to extract graph-level representations from brain functional networks, which can be used for different prediction tasks. In order to further improve the model performance, we also propose a new strategy to augment functional brain network data for contrastive learning. 
We evaluate this framework on different classification and regression tasks using the data from HCP and OASIS.
Our results from extensive experiments demonstrate the superiority of the proposed model compared to several state-of-the-art techniques.
Additionally, we use graph saliency maps, derived from these prediction tasks, to demonstrate detection and interpretation of phenotypic biomarkers.

%% file: 01_introduction.tex
\section{Introduction}

\IEEEPARstart{U}{nderstanding} brain organizations and their relationship to phenotypes (e.g., clinical outcomes, behavior or demographical variables, etc.) are of prime importance in the modern neuroscience field. 
One of important research directions is to use non-invasive neuroimaging data (e.g., functional magnetic resonance imaging or fMRI) to identify potential imaging biomarkers for clinical purposes. 
Most previous research focuses on voxel-wise and region-of-interests (ROIs) imaging features \cite{rusinek2003regional,sabuncu2015clinical,seo2015predicting}. 
However, evidences show that most of these clinical or behavior phenotypes are the outcomes of interactions among different brain regions. 
Therefore, brain networks attract more and more attention for the purpose of phenotype predictions \cite{van2012high,sporns2013human,mattar2019brain}. 
Additionally, compared to traditional neuroimaging features, brain network has more potential to gain interpretable and system-level insights into phenotype-induced brain dynamics \cite{zhang2021disentangled}. 
A brain network is a 3D brain graph model, where graph nodes represent the attributes of brain regions and graph edges represent the connections (or interactions) among these regions. 

Many studies have been conducted to analyze brain networks based on the graph theory, however, most of these studies focus on pre-defined network features, such as clustering coefficient, small-worldness \cite{beaty2018robust,brown2017prediction,eichele2008prediction,li2017predicting,warren2017brain}. 
This may be sub-optimal since these pre-defined network features may not be able to capture the characteristics of the whole brain network. 
However, the whole brain network is difficult to be analyzed due to the high dimensionality.  
To tackle this issue, Graph Neural Network (GNN), as one of embedding techniques, has gained increasing attentions to explore biological characteristics of brain network-phenotype associations in recent years \cite{hu2016clinical,kawahara2017brainnetcnn,ktena2018metric}. 
GNN is a class of deep neural networks that can embed the high-dimensional graph topological structures with graph node features into low dimensional latent space based on the information passing mechanism \cite{kipf2016semi,velivckovic2017graph,ying2019gnnexplainer}. 
A few studies proposed different GNNs to embed the nodes in brain networks and applied a global readout operation (e.g., global mean or sum) to summarize all latent node features as the whole brain network representation for downstream tasks (e.g., behavior score regression, clinical disease classification) \cite{kawahara2017brainnetcnn,ktena2018metric,zhang2019integrating}. 
However, the message passing of GNNs is inherently `flat' which only propagates information across graph edges and is unable to capture hierarchical structures rooted in graphs which are crucial in brain functional organizations \cite{ying2018hierarchical,hilgetag2020hierarchy,mastrandrea2017organization,meunier2009hierarchical}.  
To address this issue, many recent studies introduce hierarchical GNNs, including node embedding and hierarchical graph pooling strategies, to embed the whole brain network in a hierarchical manner \cite{ying2018hierarchical,lee2019self,zhang2019hierarchical,li2021braingnn,tang2021commpool}. 

Although GNNs have achieved great progresses on brain network mining, three issues should be addressed: 
\begin{itemize}
    \item Most current GNNs are designed for unsigned graphs in which all graph nodes are connected via non-negative edges (i.e., edge weights are in the range of $[0,\infty)$). 
    However, signed graphs are very common in brain research (e.g., functional MRI-derived brain networks or brain functional networks). 
    Therefore, signed graph embedding models are valuable.
    \item Brain network data, compared with other types of network data, is insufficient since the data collection is very expensive and time consuming. 
    This may limit the model performance on prediction tasks in a way. 
    \item Most current GNNs on brain network studies are not interpretable, and thus are incapable to provide biological explanations or heuristic insights for model outcomes. 
    This is mainly due to the black-box nature of the neural networks.
\end{itemize}
To tackle the first issue, a few recent studies proposed signed graph embedding models based on the balance-theory \cite{cartwright1956structural,derr2018signed,heider1946attitudes,li2020learning}. 
The balance-theory, motivated by human attitudes in social networks, is used to describe the node relationship in signed graphs, where nodes connected by positive edges are considered as `friends' otherwise are considered as `opponents'. 
Meanwhile, the balance-theory also defines $4$ higher-order relationships among graph nodes: (1) the `friend' of `friend' is `friend', (2) the `opponent' of `friend' is `opponent', (3) the `friend' of `opponent' is `opponent' and (4) the `opponent' of `opponent' is `friend'.
These definitions are accorded with nodal relationships in the functional brain network, which indicates that the balance theory might be applicable in functional brain network embedding.  
However, existing signed graph embedding models focus on embedding graph nodes with signed edges into latent features without considering the hierarchical structures in graphs, which may not facilitate the whole graph representation learning and the graph-level tasks (i.e., clinical disease classification based on whole brain networks).
To address this issue, we propose a hierarchical graph pooling module on signed graphs based on the information theory and extend the current methods to a hierarchical signed graph embedding model. 

To address the second issue, we propose a data augmentation strategy to augment functional brain networks. 
Meanwhile, we introduce the graph contrastive learning architecture, where contrastive graph samples are generated by the proposed augmentation strategy, to boost the model performance on prediction tasks.
The data augmentation aims at creating reasonable data samples, by applying certain transformations, which are similar to the original ones. 
For example, image rotation and cropping are common transformations to generate new samples in image classification tasks \cite{khosla2020supervised,berthelot2019mixmatch,xie2020unsupervised}. 
In graph structural data, a few studies proposed to utilize graph perturbations (i.e., add/drop graph nodes, manipulate graph edges) and graph view augmentation (e.g., graph diffusion) to generate contrastive graph samples from different views \cite{hassani2020contrastive,you2020graph,zhu2021graph,zhao2020datac}.
These strategies, although boosting the model performance on large-scale benchmark datasets (e.g., CORA, CITESEER, etc.), may not be suitable to generate contrastive brain network samples.
On the one hand, each node in brain networks represents a defined brain region with specific brain activity information so that the brain node can not be arbitrarily removed or added. 
On the other hand, add/drop operations on brain network may lead to unexpected model outcomes which are difficult to explain and understand from biological views.
Therefore, we generate the augmented brain functional networks directly from fMRI BOLD signals, where the generated samples are similar and the biological structure is maintained.

As for the last issue, our proposed graph pooling module is interpretable by nature. 
Previous studies indicated that brain networks are hierarchically organized by some regions as neuro-information hubs and peripheral regions, respectively \cite{van2013network,ilyas2011distributed,hwang2013development}.
Within our graph pooling module, an information score is designed to measure the information gain for each brain node and only top-$K$ nodes with high information gains will be preserved as brain information hubs while the information of other peripheral brain nodes will be aggregated onto these hubs.
Hence, the proposed pooling module can be interpreted as a brain information hubs generator. 
Apparently, the outcome of this pooling module is a subgraph of the original brain network without any new nodes. 
Therefore, yielded subgraph nodes can be regarded as potential biomarkers to provide heuristic biological explanations for tasks. 

Our main contributions are summarized as follow:
\begin{itemize}
    \item We propose a \underline{h}ierarchical \underline{s}igned \underline{g}raph \underline{r}epresentation \underline{l}earning (HSGRL) model to embed brain functional networks and we apply the proposed model on multiple phenotype prediction tasks. 
    \item We propose an augmentation strategy for fMRI-derived brain network data. 
    To further boost the model performance, we build up a contrastive learning framework with the proposed HSGPL model, where the contrastive samples are generated by the designed augmentation strategy. 
    \item The proposed HSGPL model is interpretable which yields heuristic biological explanations. 
    \item Extensive experiments are conducted to demonstrate the superiority of our method. 
    Moreover, we draw graph saliency maps for clinical tasks, to enable interpretable detection of phenotype biomarkers. 
\end{itemize}


%% file: 02_relatedwork.tex
\section{Related Works}
\subsection{Graph Neural Networks and Brain Network Embedding}
GNNs are generalized deep learning architectures which are broadly utilized for graph representation learning in many fields (e.g., social network mining \cite{chen2018fastgcn,huang2018adaptive}, molecule studies \cite{dai2016discriminative,duvenaud2015convolutional} and brain network analysis \cite{liu2019community}).
Most existing GNN models (e.g., GCN \cite{kipf2016semi}, GAT \cite{velivckovic2017graph}, GraphSage \cite{hamilton2017inductive}) focus on node-level representation learning and only propagate information across edges of the graph in a flat way.
When deploying these models on graph-level tasks (e.g., graph classification, graph similarity learning, \cite{li2015gated,vinyals2015order,zhang2018end,ma2021deep}), the whole graph representations are obtained by a naive global readout operation (e.g., sum or average all node feature vectors). 
However, this may lead to poor performance and low efficiency in graph-level tasks since the hierarchical structure, an important property that existed in graphs, is ignored in these models. 
To explore and capture hierarchical structures in graphs, a few hierarchical graph pooling strategies are proposed to learn representations for the whole graph in a hierarchical manner \cite{lee2019self,ying2018hierarchical,gao2019graph,yuan2020structpool,zhang2019hierarchical}.  
Traditional methods to extract brain network patterns are based on graph theory \cite{beaty2018robust,brown2017prediction,eichele2008prediction,li2017predicting,warren2017brain} or geometric network optimization \cite{korthauer2018disrupted,zhan2015boosting,cao2017t,sui2011discriminating}.
A few recent studies \cite{kawahara2017brainnetcnn,ktena2018metric,zhang2019new} introduce GNNs to discover brain patterns for phenotypes predictions. 
However, hierarchical structures in brain networks are not considered in these models, which limits the model performance in a way. Recently, a few hierarchical brain network embedding models are proposed\cite{jiang2020hi,li2021braingnn}. 

However, all the aforementioned GNNs are designed for unsigned graph representation learning.
A few recent studies are proposed to handle the signed graphs, however, they only consider the node-level representation learning \cite{derr2018signed,li2020learning,jung2020signed,shen2018deep}.
In this work, we design a signed graph hierarchical pooling strategy to extract graph-level representations from brain functional networks.   

\subsection{Interpretable Graph Learning Model}
Generally, the mechanism about how GNNs embed the graph nodes can be explained as a message passing process, which includes message aggregations from neighbor nodes and message (non-linear) transformations \cite{ying2019gnnexplainer,huang2020graphlime,li2021braingnn}.   
However, most current hierarchical pooling strategies are not interpretable \cite{ying2018hierarchical,zhang2019hierarchical,lee2019self}.
A few recent studies try to propose interpretable graph pooling strategies to make the pooling module intelligible to the model users.
Most of these pooling strategies down-sample graphs relying on network communities which are one of the important hierarchical structures that can be interpreted \cite{tang2021commpool,li2021braingnn,cui2021brainnnexplainer}.
For example, \cite{li2021braingnn} proposed a hierarchical graph pooling neural network relying on brain network community to yield interpretable biomarkers. 
The hierarchical pooling strategy proposed in this work relies on the network information hub which is another important hierarchical structure in brain networks. 

\subsection{Data Augmentation for Graph Contrastive Learning} 
Most current graph contrastive learning methods augment graph contrastive samples by manipulating graph topological structures. 
For example, \cite{you2020graph,zhu2021graph} generate the contrastive graph samples by dropping nodes and perturbing edges. 
Other studies generate contrastive samples by changing the graph local receptive field, which is named as the graph view augmentation \cite{xu2021infogcl,hassani2020contrastive}.
In this work, we introduce the graph contrastive learning into brain functional network analysis and generate contrastive samples from the fMRI BOLD signals.

%% file: 03_prelim.tex
\section{Preliminaries of Brain Functional Networks}
We denote a brain functional network with $N$ nodes as $G=\{V, E\}=(A, H)$. 
$V$ is the graph node set where each node (i.e., $v_{i}, i=1, ..., N$) represents a brain region. 
$E$ is the graph edge set where each edge (i.e., $e_{i,j}$) describes the connection between node $v_{i}$ and $v{j}$. 
$A \in \mathbb{R}^{N \times N}$ is the graph adjacency matrix where each element, $a_{i, j} \in A$, is the weight of edge $e_{i,j}$. 
$H \in \mathbb{R}^{N \times C}$ is the node feature matrix where $H_{i} \in H$ is the $i-th$ row of $H$ representing the feature vector of $v_{i}$. 
Let $B \in \mathbb{R}^{N \times D}$ be the fMRI BOLD signal matrix, where $D$ is the signal length. 
Generally, the edge weight in the brain functional network can be computed from the fMRI BOLD signal by $ a_{i, j} = corr (b_{i}, b_{j})$, where $b_{i}$ is the $i-th$ row of $B$ representing the BOLD signal of $v_{i}$ and $corr(\cdot)$ is the correlation coefficient operator. 
Note that $a_{i,j}$ can be either positive or negative value so that brain functional network is a signed graph. 
For each subject, we use $\hat{}$ and $\check{}$ to denote a functional brain network contrastive sample pair (i.e., $[\hat{G}=(\hat{A}, \hat{H}), \check{G}=(\check{A}, \check{H})]$). 

%% file: 04_method.tex
\section{Methodology}
\begin{figure*}[t]
\centering
\includegraphics[width=1.0\textwidth]{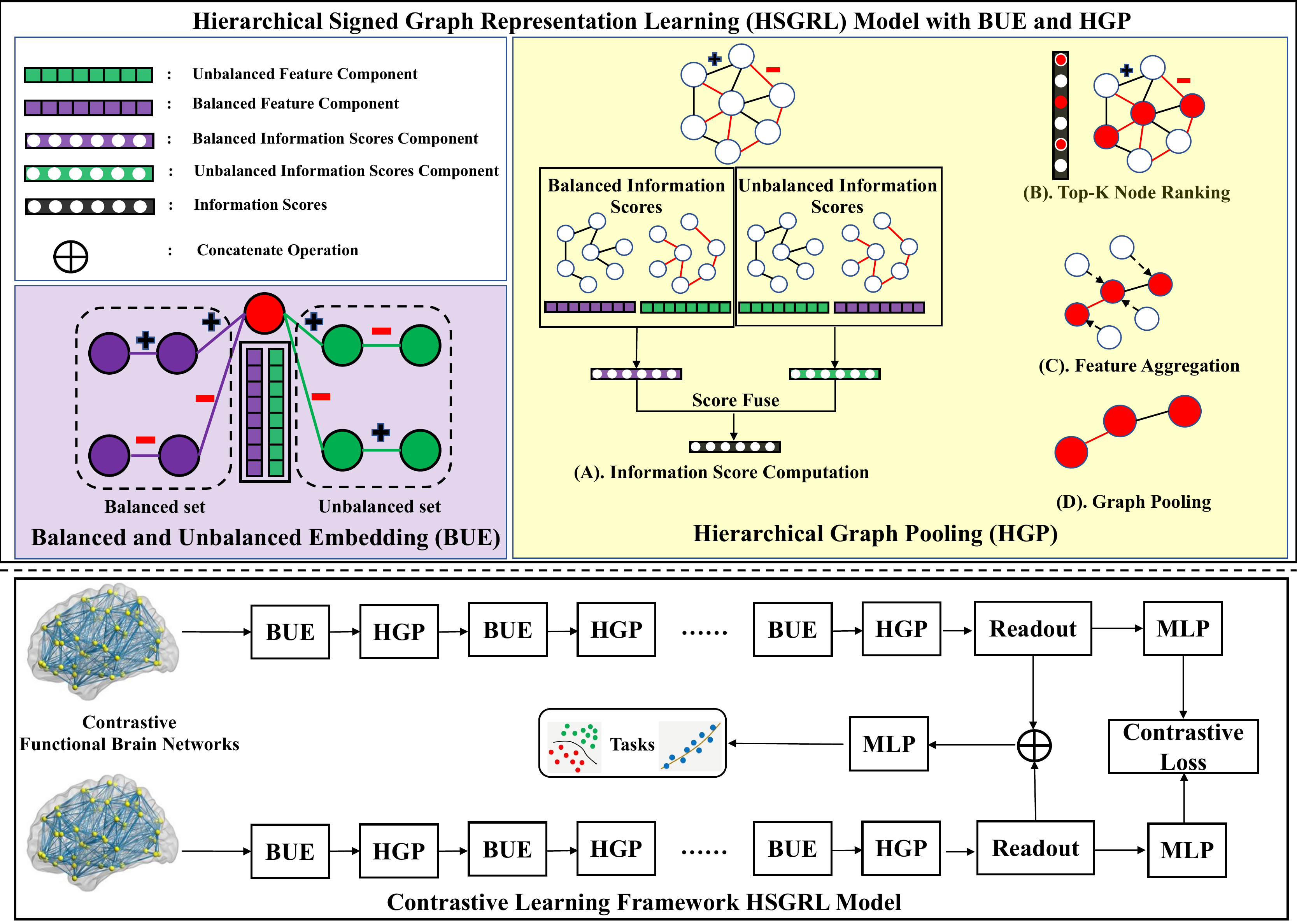}
\caption{Diagram of the proposed contrastive graph learning framework (in the bottom black box) with hierarchical signed graph representation learning model (in the top black box) for functional brain network embedding and downstream tasks (i.e., phenotype classification or regression).}
\label{main_figure}
\end{figure*}

In this section, we first propose a data augmentation strategy to generate contrastive samples for brain functional networks. 
Secondly, we introduce our proposed hierarchical signed graph representation learning (HSGRL) model with node embedding and hierarchical graph pooling modules. 
Finally, we deploy the contrastive learning framework on our proposed HSGRL model to yield the representations for the whole graph, which can be applied to downstream prediction tasks.    

\subsection{Contrastive Samples of Brain Functional Networks}
The generation of contrastive samples aims at creating reasonable and similar functional brain network pairs by applying certain transformations. 
Here we propose a new strategy to generate the brain functional network contrastive samples from fMRI BOLD signals. 
For each node $v_{i}$, we generate two sub-BOLD-signals ($\hat{b}_{i}$ and $\check{b}_{i}$) by manipulating its original bold signal $b_{i}$.
Specifically, we use a window ($size = d$) to clamp the $b_{i}$ from the signal head and tail, respectively:
\begin{eqnarray}
    \hat{b}_{i}   &=& b_{i}[d+1, \quad d+2, \quad ..., \quad D]  \nonumber \\
    \check{b}_{i} &=& b_{i}[1, \quad 2, \quad ..., \quad D-d]
\end{eqnarray}
Obviously, $b_{i} \in \mathbb{R}^{1 \times D}$, $\hat{b}_{i}$ and $\check{b}_{i} \in \mathbb{R}^{1 \times (D-d)}$. 
To keep the similarity between $\hat{G}$ and $\check{G}$, we set the window size $d \ll D$. 
After we generate a pair of sub-bold-signals, we can compute edge weights of the pairwise contrastive brain functional network samples by:
\begin{eqnarray}
    \hat{a}_{i, j} &=& corr (\hat{b}_{i}, \hat{b}_{j})  \nonumber \\
    \check{a}_{i, j} &=& corr (\check{b}_{i}, \check{b}_{j}),  
\end{eqnarray}
where $\hat{a}_{i, j} \in \hat{A}$ and $\check{a}_{i, j} \in \check{A}$ are the weights of $e_{i, j}$ in two contrastive samples. 
We do not consider the contrastive node features in this work, therefore $\hat{X} = \check{X} = X$.  
The generated contrastive sample pairs are similar with same node features and slightly different edge weights. 
We will show this similarity in section \ref{sim_contrastive_sample_section}.


\subsection{Hierarchical Signed Graph Representation Learning Model}
We present our Hierarchical Signed Graph Representation Learning (HSGRL) model in Fig. \ref{main_figure}. 
The HSGRL model includes Balanced and Unbalanced Embedding (BUE) module and Hierarchical Graph Pooling (HGP) module. 

\subsubsection{BUE module}
The balance theory is broadly used to analyze the node relationships in signed graphs. 
The theory states that given a node $v_{i}$ in a signed graph, any other node (i.e., $v_{j}$) can be assigned into either balanced node set or unbalanced node set to $v_{i}$ regarding to a path between $v_{i}$ and $v_{j}$. 
Specifically, if the number of negative edges are even in the path between $v_{i}$ and $v_{j}$, then $v_{j}$ belongs to the balanced set of $v_{i}$. 
Otherwise, $v_{j}$ belongs to the unbalanced set of $v_{i}$. 
The balance theory indicates that:
\begin{itemize}
    \item Each graph node, $v_{j}$, can belong to either the balanced or unbalanced node set of a given target node $v_{i}$. 
    \item The path between $v_{i}$ and $v_{j}$ determines the balance attribute of $v_{j}$. 
\end{itemize}
Motivated by this, we adopt the idea of signed graph attention networks from \cite{li2020learning} to embed brain functional network nodes to generate latent node features with balanced and unbalanced components: 
\begin{eqnarray}
    X^{B}, X^{U} = F_{sign} (A, H)
\end{eqnarray}
where $F_{sign}(\cdot)$ is the signed graph attention encoder \cite{li2020learning}.
$X^{B}$ and $X^{U}$ are the node balanced and unbalanced components of node latent features, respectively. 
We fuse the two feature components as the node latent features by:
\begin{eqnarray}
X = [X^{B} \| X^{U}],
\end{eqnarray}
where $[||]$ denotes concatenate operation.



\subsubsection{Hierarchical Signed Graph Pooling}
As shown in Fig \ref{main_figure}, the proposed Hierarchical Graph Pooling (HGP) module consists of $4$ steps including: (A) information scores computation, (B) Top-K informative hubs selection, (C) features aggregation and (D) graph pooling.

\textbf{\textit{Information Score Computation:}}
The information score of each node is also considered to contain balanced and unbalanced components to measure the information quantity that each node gains from balanced node set and unbalanced node set, respectively. 
We first split the signed graph (i.e., with adjacency matrix as $A$) into positive sub-graph (with adjacency matrix as $A_{+}$) and negative one (with adjacency matrix as $A_{-}$). 
Then we utilize Laplace normalization to normalize these two adjacency matrices as:
\begin{eqnarray}
\Bar{A}_{+} &=& D^{-\frac{1}{2}}_{+} A_{+} D^{-\frac{1}{2}}_{+} \nonumber \\
\Bar{A}_{-} &=& D^{-\frac{1}{2}}_{-} \lvert A_{-} \rvert D^{-\frac{1}{2}}_{-},
\end{eqnarray}
where $\Bar{A}$ is the normalized adjacency matrix. $D_{+}$ and $D_{-}$ are degree matrices of $A_{+}$ and $| A_{-} |$, respectively. 
Note that the i-th line in $\Bar{A}$, denoted by $\Bar{A}_{i}$, represents the connectivity probability distribution between $v_{i}$ and any other nodes. 
For each node (i.e., $v_{i}$), we respectively define the balanced and unbalanced components of information score (IS) by:
\begin{eqnarray}
IS^{B}_{i} &=& \left\| \Bar{A}_{+,i:}^\top \otimes X^{B} \right\|_{\Tilde{L}_{1}} + \left\| \Bar{A}_{-,i:}^\top \otimes X^{U} \right\|_{\Tilde{L}_{1}} \nonumber \\
IS^{U}_{i} &=& \left\| \Bar{A}_{+,i:}^\top \otimes X^{U} \right\|_{\Tilde{L}_{1}} + \left\| \Bar{A}_{-,i:}^\top \otimes X^{B} \right\|_{\Tilde{L}_{1}},
\end{eqnarray}
where $\left \| \cdot \right \|_{\Tilde{L}_{1}}$ is line-wise $L_{1}$ norm, and $\otimes$ is the scalar-multiplication between each line of two matrices.
$\top$ represents transpose of vector.
Then the IS of $v_{i}$ can be obtained by:
\begin{eqnarray}
IS_{i} = IS^{B}_{i} + IS^{U}_{i}.
\end{eqnarray}

\textbf{\textit{Top-$K$ Node Selection and Feature Aggregation:}}
After we obtain the information score for each brain node, we rank the IS and select $K$ brain nodes, with top-$K$ IS values, as informative network hubs.
For the other nodes, we aggregate their features on the selected $K$ network hubs based on the feature attention. 
Particularly, the feature attention between $v_{i}$ and $v_{j}$ is computed by: $x_{i}x_{j}^\top$.
We weighted add (i.e., set feature attentions as weights) the feature of each unselected node to one of hub features, where the attention value between these two nodes is the biggest.  

\textbf{\textit{Graph Pooling}}
After the feature aggregation, we down-scale the graph node by removing all unselected nodes. 
In another word, only the selected top-$K$ network hubs as well as the edges among them will be preserved after graph pooling. 
Since the functional brain network is a fully connected graph so that no isolated node is existed in the down-scaled graph.  

\subsection{Contrastive Learning Framework with BUE and HGP}
The contrastive learning framework with HSGRL is presented in Fig. \ref{main_figure}. 
Assume that we forward a pair of contrastive graph samples into the proposed HSGRL model, we will obtain two node latent features, $\hat{X}$ and $\check{X}$ after the last pooling module. 
We first generate the graph-level representations of two functional brain networks based on the latent node features by a readout operator:
\begin{eqnarray}
\hat{X}_{G} = \sum_{i=1}^{N^\prime} \hat{x}_{i}, \quad
\check{X}_{G} = \sum_{i=1}^{N^\prime} \check{x}_{i},
\end{eqnarray}
where $\hat{x}_{i}$ and $\check{x}_{i}$ are $i-th$ row of $\hat{X}$ and $\check{X}$.
$N^\prime(<N)$ is the number of nodes in the down-scaled graph generated by the last pooling module. 

\subsubsection{Contrastive Loss}
The normalized temperature-scaled cross entropy loss \cite{wu2018unsupervised,sohn2016improved,van2018representation} is utilized to construct the contrastive loss. 
In the framework training stage, we randomly sample $M$ pairs from the generated contrastive graph samples as a mini-batch and forward them to the proposed HSGRL model to generate contrastive graph representation pairs (i.e., $\hat{X}_{G}$ and $\check{X}_{G}$).
We use $m \in \{1,..., M\}$ to denote the ID of the sample pair. 
The contrastive loss of the $m-th$ sample pair is fomulated as:
\begin{eqnarray}
\ell_{m} = -log \frac{exp(\Phi(\hat{X}_{G}^{m}, \check{X}_{G}^{m})/\alpha)}
{\sum_{t=1, t \neq m}^{M} exp(\Phi (\hat{X}_{G}^{m}, \check{X}_{G}^{t})/\alpha)},
\end{eqnarray}
where $\alpha$ is the temperature parameter. $\Phi(\cdot)$ denotes a similarity function that:
\begin{equation}
  \Phi(\hat{X}_{G}^{m}, \check{X}_{G}^{m}) = \hat{X}_{G}^{m \top} \check{X}_{G}^{m} / \|\hat{X}_{G}^{m}\| \|\check{X}_{G}^{m}\|.  
\end{equation}
The batch contrastive loss can be computed by:
\begin{equation}
    \mathcal{L}_{contrastive} = \frac{1}{M}\sum_{m=1}^{M} \ell_{m}
\end{equation}

\subsubsection{Downstream Task and Loss Functions}
We use an MLP to generate the framework prediction for both classification and regression tasks. 
Specifically, the prediction can be generate by $ Y_{pred} = MLP ([\hat{X}_{G} \| \check{X}_{G}])$.
We use $NLLLoss$ and $L_{1}Loss$ as supervised loss functions ($\mathcal{L}_{supervised}$) of classification and regression tasks, respectively. 
The whole framework can be trained in an end-to-end manner by optimizing:
\begin{equation}
    \mathcal{L} = \mu_{1} \mathcal{L}_{supervised} + \mu_{2} \mathcal{L}_{contrastive},
\end{equation}
where $\mu_{1}$ and $\mu_{2}$ are the loss weights.

%% file: 05_experiment.tex
\begin{figure}[htp]
\centering
\includegraphics[width=0.45\textwidth]{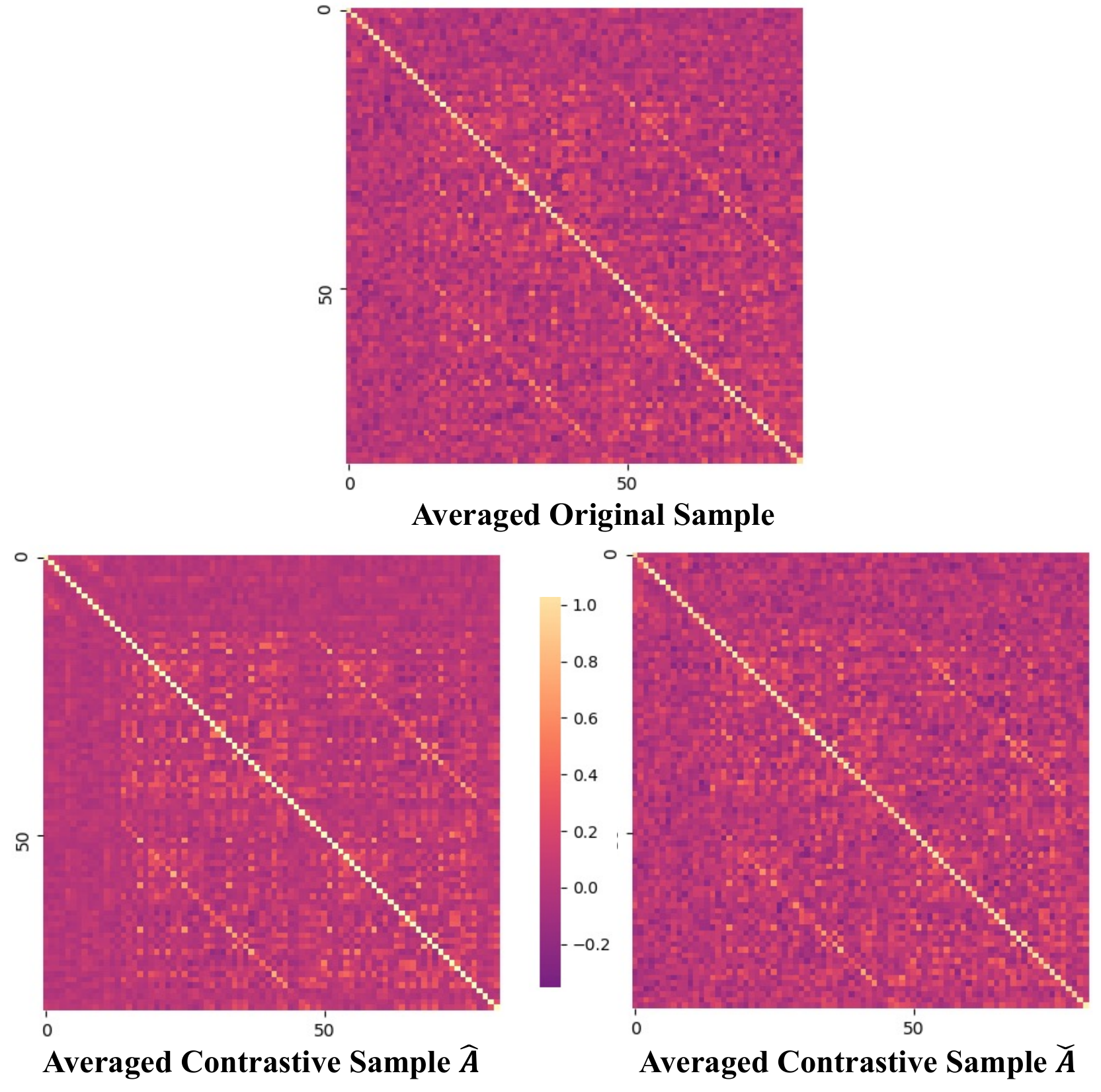}
\caption{Visualization of the averaged adjacency matrices for original and contrastive samples. The averaged contrastive sample pair is generated by using a window size $d=10$.}
\label{contrast}
\end{figure}

\section{Experiments}
\label{experiments}

\subsection{Datasets and Data Preprocessing}
Two publicly available datasets were used to evaluate our framework.
The first includes $1206$ young healthy subjects (mean age $28.19 \pm 7.15$, $657$ women) from the Human Connectome Project (HCP) \cite{van2013wu}. 
The second includes 1326 subjects (mean age $=70.42 \pm 8.95$, $738$ women) from the Open Access Series of Imaging Studies (OASIS) dataset \cite{lamontagne2019oasis}. 
Details of each dataset can be found on their official websites 
\footnote{\url{https://www.oasis-brains.org}} \footnote{\url{https://wiki.humanconnectome.org}}. 
CONN~\cite{whitfield2012conn} were used to preprocess fMRI data and the preprocessing pipeline follows our previous publications~\cite{fortel2020connectome,ajilore2013constructing}. For HCP data, each subject's network has a dimension of $82 \times 82$ based on 82 ROIs defined using FreeSurfer (V6.0)~\cite{fischl2012freesurfer}. 
For OASIS data, each subject's network has a dimension of $132 \times 132$ based on the Harvard-Oxford Atlas and AAL Atlas. 
We deliberately chose different network resolutions for HCP and OASIS to evaluate whether the performance of our new framework is affected by the network dimension or atlas. 
\subsection{Implementation Details}
We randomly split the entire functional brain network dataset into $5$ disjoint subsets for $5$-fold cross-validations in our experiments. 
The values in the adjacency matrices ($\hat{A}$ and $\check{A}$) of brain functional networks are within range of $[-1, 1]$. 
We compute the kurtosis and skewness values of the fMRI BOLD signals as the node feature matrices ($H$). 
We use the Adam optimizer \cite{kingma2014adam} to optimize the loss functions in our model with a batch size of $128$. 
The initial learning rate is $1e^{-4}$ and decayed by $(1-\frac{current\_epoch}{max\_epoch})^{0.9}$.
We also regularized the training with an $L_{2}$ weight decay of $1e^{-5}$.
We set the maximum number of training epochs as $1000$ and, following the strategy in \cite{lee2019self},\cite{shchur2018pitfalls}, stop training if the validation loss does not decrease for $50$ epochs. 
The experiments were deployed on one NVIDIA RTX A6000 GPU. 
\subsection{Similarities of Contrastive Samples}
\label{sim_contrastive_sample_section}
We utilize the $L_{2}$ distance and Cosine Similarity to measure the similarities of the adjacency matrices of contrastive brain networks.
Here, we set the window size $d=10$ to generate the contrastive adjacency matrices. 
The inner-pair similarity is computed by $\frac{1}{M} \sum_{m=1}^{M} \Psi(\hat{A}^{m}, \check{A}^{m})$, and the inter-pair similarity is computed by $\frac{1}{M^{2}} \sum_{m=1}^{M}\sum_{t=1}^{M} \Psi(\hat{A}^{m}, \check{A}^{t})$, where $\Psi(\cdot)$ is the similarity function (i.e., $L_{2}$ distance or Cosine Similarity). 
The inner-pair  $L_{1}$ distances on HCP and OASIS data are $0.1301$ and $0.0915$, respectively. 
The inner-pair Cosine Similarities on HCP and OASIS data are $0.9283$ and $0.9466$, respectively. 
The inter-pair  $L_{1}$ distances on HCP and OASIS data are $0.2925$ and $0.3137$, respectively. 
The inter-pair Cosine Similarities on HCP and OASIS data are $0.7311$ and $0.7014$, respectively.
We visualize the averaged adjacency matrics on HCP data in Fig. \ref{contrast} to show their similarities. 
The original sample is generated by using the whole fMRI BOLD signal (i.e., $d=0$).    
\subsection{Classification Tasks}
\label{class-task}
\begin{table*}[t]
\centering
\caption{Classification accuracy with s.t.d values under 5-fold cross-validation on gender classification, zygosity classification and AD classification tasks. The values in \textbf{bold} show the best results.}
\begin{tabular}{l|lllll|lll}
\hline
\multicolumn{1}{c|}{\multirow{3}{*}{\textbf{Method}}} & \multicolumn{5}{c|}{\textbf{HCP}}                                                                                                                           & \multicolumn{3}{c}{\textbf{OASIS}}                                            \\ \cline{2-9} 
\multicolumn{1}{c|}{}                                 & \multicolumn{3}{c|}{Gender}                                                                     & \multicolumn{2}{c|}{Zygosity}                             & \multicolumn{3}{c}{AD}                                                        \\ \cline{2-9} 
\multicolumn{1}{c|}{}                                 & \multicolumn{1}{c}{Acc.} & \multicolumn{1}{c}{Pre.} & \multicolumn{1}{c|}{F1.}                  & \multicolumn{1}{c}{Acc.} & \multicolumn{1}{c|}{Macro-F1.} & \multicolumn{1}{c}{Acc.} & \multicolumn{1}{c}{Pre.} & \multicolumn{1}{c}{F1.} \\ \hline
t-BNE                                                 & 63.84(2.09)              & 64.17(1.90)              & \multicolumn{1}{l|}{63.264(2.12)}         & 37.19(2.65)              & 39.67(3.04)                    & 61.26(2.31)              & 63.58(2.06)              & 62.05(1.97)             \\
mCCA-ICA                                              & 61.21(4.03)              & 63.11(3.75)              & \multicolumn{1}{l|}{62.20(3.59)}          & 35.51(4.64)              & 38.71(3.34)                    & 63.37(1.98)              & 62.06(2.12)              & 64.37(2.09)             \\ \hline
SAGPOOL                                               & 68.12(3.07)              & 69.96(2.48)              & \multicolumn{1}{l|}{67.51(2.65)}          & 49.91(2.22)              & 51.07(2.31)                    & 67.23(2.15)              & 68.83(1.13)              & 67.51(2.51)             \\
DIFFPOOL                                              & 72.06(2.28)              & 74.05(1.90)              & \multicolumn{1}{l|}{73.07(2.42)}          & 53.37(1.88)              & 54.28(2.14)                    & 72.79(1.66)              & 71.55(2.15)              & 70.83(2.01)             \\
BrainCheby                                            & 75.08(1.98)              & 76.14(2.38)              & \multicolumn{1}{l|}{74.09(1.84)}          & 56.25(2.12)              & 57.37(2.05)                    & 72.55(2.45)              & 73.36(1.88)              & 72.62(1.33)             \\
BrainNet-CNN                                          & 74.09(2.49)              & 73.71(1.96)              & \multicolumn{1}{l|}{73.27(2.21)}          & 54.03(2.20)              & 55.25(2.46)                    & 68.37(1.71)              & 69.97(1.30)              & 68.51(2.02)             \\ \hline
Ours w/o Contra.                                      & \textbf{78.86(2.18)}     & \textbf{80.06(1.33)}     & \multicolumn{1}{l|}{\textbf{77.52(1.69)}} & \textbf{61.05(1.70)}     & \textbf{63.24(2.51)}           & \textbf{76.26(2.32)}     & \textbf{75.42(1.62)}     & \textbf{76.80(1.72)}    \\
Ours                                                  & \textbf{81.51(1.14)}     & \textbf{82.37(1.95)}     & \multicolumn{1}{l|}{\textbf{80.69(2.03)}} & \textbf{63.33(2.06)}     & \textbf{64.51(1.74)}           & \textbf{77.51(1.84)}     & \textbf{78.83(1.78)}     & \textbf{78.28(1.95)}    \\ \hline
\end{tabular}
\label{classification}
\end{table*}

\subsubsection{Experiment Setup} 
Six baseline models are utilized for comparison, including two machine learning graph embedding models (t-BNE \cite{cao2017t} and mCCA-ICA \cite{sui2011discriminating}), two deep graph representation learning models designed for brain network embedding (BrainChey \cite{ktena2018metric} and BrainNet-CNN \cite{kawahara2017brainnetcnn}), and two hierarchical graph neural networks with graph pooling strategies (DIFFPOOL \cite{ying2018hierarchical} and SAGPOOL \cite{lee2019self}).
Meanwhile, we compare our model with and without optimizing contrastive loss to show that the contrastive learning is beyond the data augmentation.    
The results for gender and Alzheimer Disease (AD) classification are reported in accuracy, precision and F1-score with their standard deviation (\emph{std}). 
The results for zygosity classification (i.e., $3$ classes classification task with class labels as: not twins, monozygotic twins and dizygotic twins) are reported in accuracy and Macro-F1-score with their \emph{std}. 
The number of BUE and HGP modules are set to $3$. 
We search the loss weights $\mu_{1}$ and $\mu_{2}$ in range of $[0.1, 1, 5]$ and $[0.01, 0.1, 0.5, 1]$ respectively and determine the loss weights as $\mu_{1} = 1$, $\mu_{2} = 0.1$. 
The temperature parameter in contrastive loss is set as $0.2$. 
Details of the hyperparameters analysis are shown in section \ref{abalation-studies}.

\subsubsection{Results}
Table \ref{classification} shows the results of gender classification, zygosity classification and AD classification. 
It shows that our model achieves the best performance comparing to all baseline methods on three tasks. 
For example, in the gender classification, our model outperforms the baselines with at least $8.56\%$, $8.18\%$ and $8.91\%$ increases in accuracy, precision and F1 scores, respectively. 
In general, the deep graph neural networks are superior than the traditional graph embedding methods (i.e., t-BNE and mCCA-ICA). 
When we remove the supervision of the contrastive loss, the performance, though comparable to baselines, decreases in a way. 
This manifests the effectiveness of the contrastive learning and indicates that the contrastive learning is beyond a data augmentation strategy.

\begin{figure*}[h]
\small
    \centering
    \raggedright
    \begin{minipage}[h]{1.0\linewidth}
    \centering
    \includegraphics[width=6in]{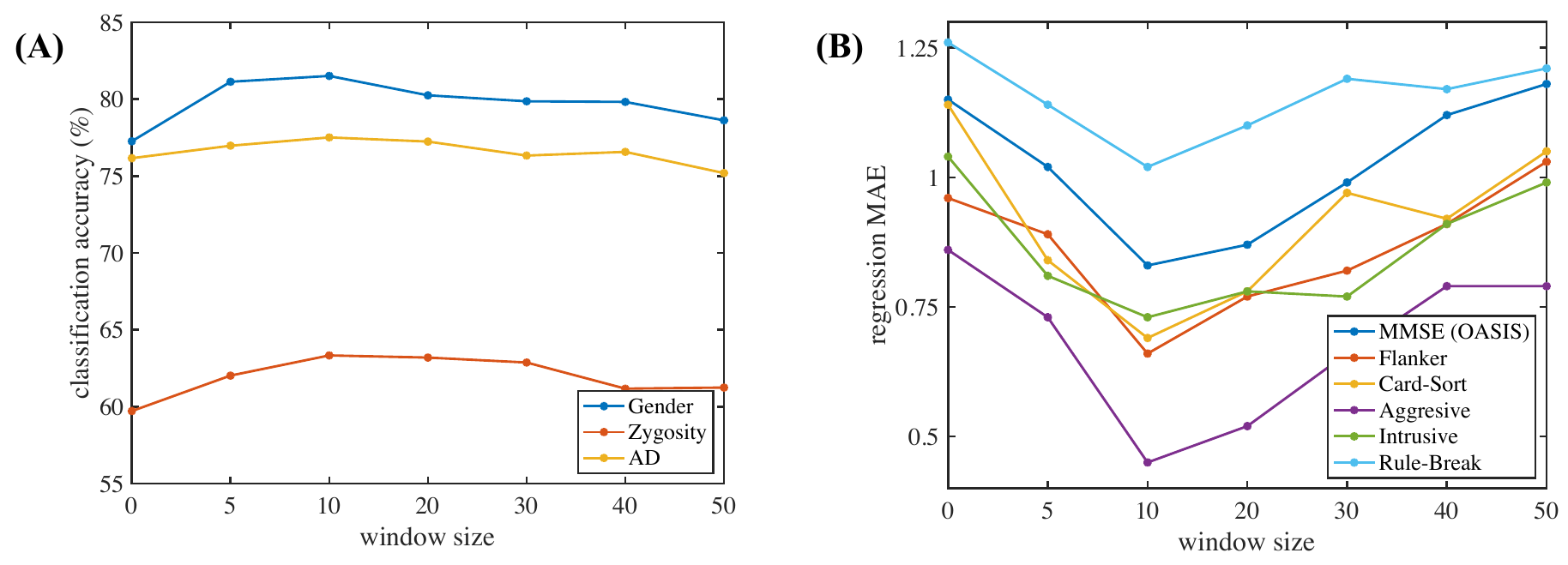}
    \end{minipage}
    \caption{The model performance obtained with different contrastive samples generated by different window sizes. \textbf{(A)} shows the analysis on classification tasks and \textbf{(B)} shows the analysis on regression tasks.}
    \label{para_win}
\end{figure*}
\begin{figure*}[h]
\small
    \centering
    \raggedright
    \begin{minipage}[h]{1.0\linewidth}
    \centering
    \includegraphics[width=7in]{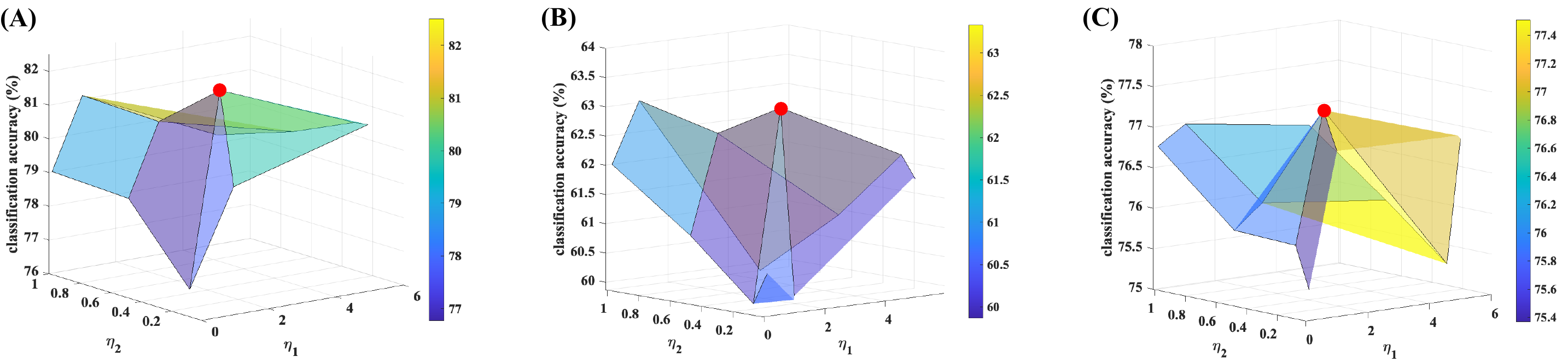}
    \end{minipage}
    \caption{Loss weights analysis on classification tasks. \textbf{(A)} shows the analysis on gender classification, \textbf{(B)} shows the analysis on zygosity classification and \textbf{(C)} shows the analysis on AD classification. The red points represent the best results.}
    \label{para_class_loss}
\end{figure*}
\begin{figure*}[h]
\centering
\includegraphics[width=1.0\textwidth]{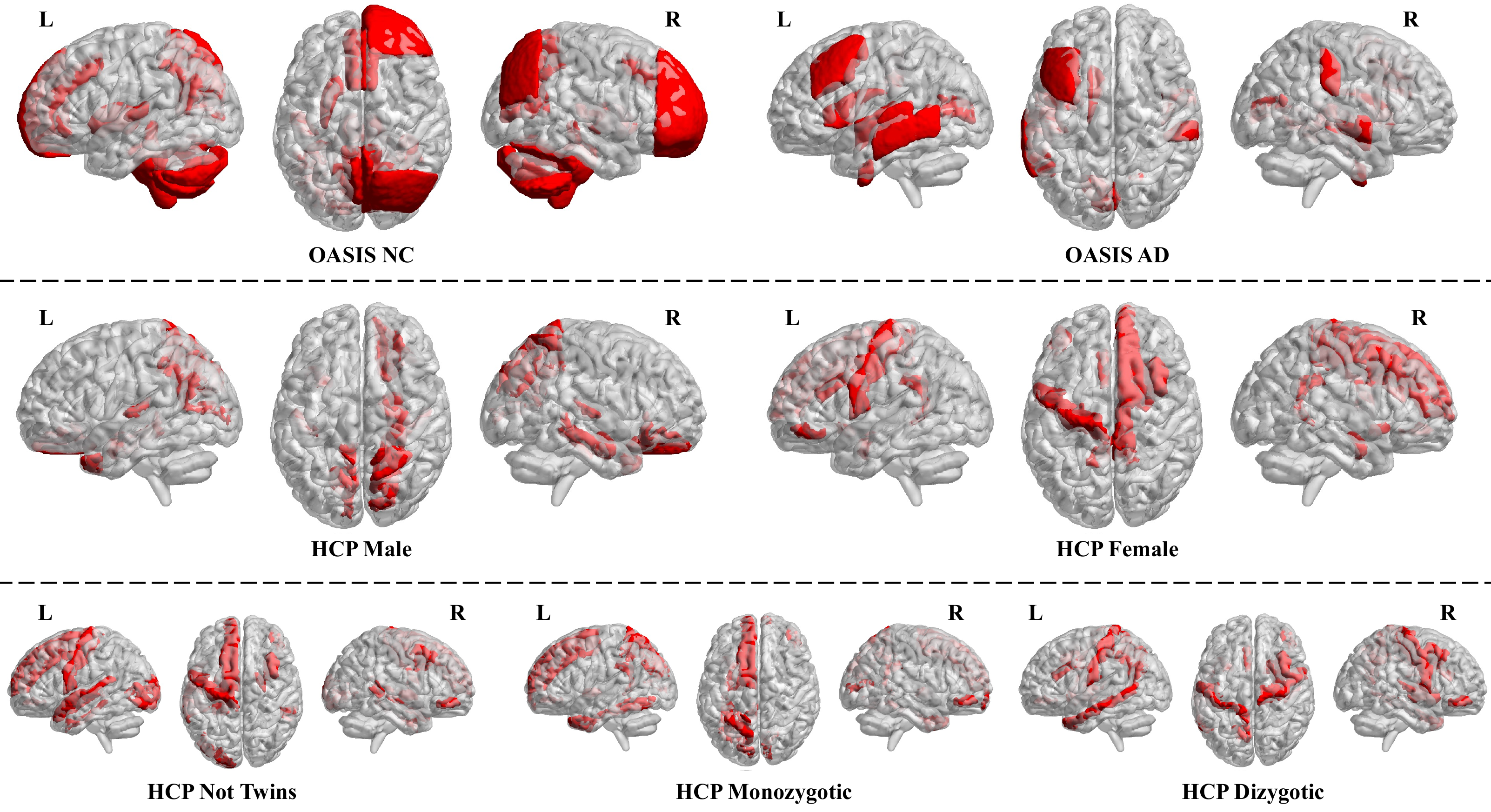}
\caption{Brain saliency maps for classification tasks. Here we identify: (1) top 15 regions associated with AD and NC from OASIS, (2) top 10 regions associated with each sex and each zygosity from HCP.}
\label{SaliencyMaps_classification}
\end{figure*}
\begin{figure*}[h]
\centering
\includegraphics[width=1.0\textwidth]{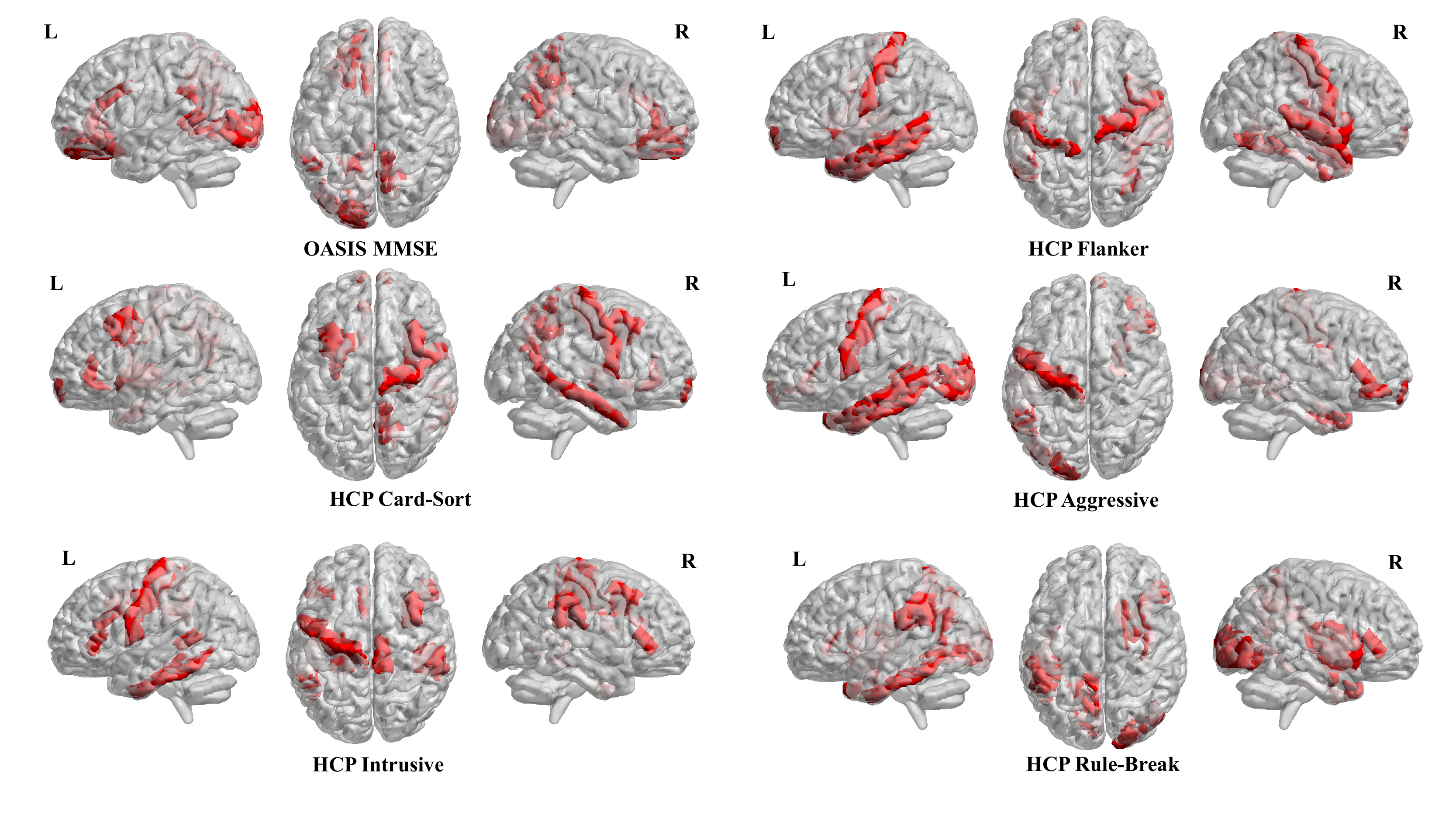}
\caption{Brain saliency maps for regression tasks. Here we identify: (1) top 15 regions associated with MMSE from OASIS, (2) top 10 regions associated with Flanker score, Card-Sort score, Aggressive score, Intrusive score and Rule-Break score from HCP.}
\label{SalienceMaps_regression}
\end{figure*}
\subsection{Regression tasks}
\label{regression-task}
\begin{table*}[t]
\centering
\caption{Regression Mean Absolute Error (MAE) with s.t.d under 5-fold cross-validation. The values in \textbf{bold} show the best results.}
\begin{tabular}{l|l|lllll}
\hline
\multicolumn{1}{c|}{\multirow{2}{*}{\textbf{Method}}} & \multicolumn{1}{c|}{\textbf{OASIS}} & \multicolumn{5}{c}{\textbf{HCP}}                                                                                                                                                         \\ \cline{2-7} 
\multicolumn{1}{c|}{}                                 & \multicolumn{1}{c|}{MMSE}            & \multicolumn{1}{c}{Flanker} & \multicolumn{1}{c}{Card-Sort} & \multicolumn{1}{c}{Aggressive} & \multicolumn{1}{c}{Intrusive} & \multicolumn{1}{c}{Rule-Break} \\ \hline
t-BNE                                                 & 2.02(0.36)                                        & 1.69(0.19)                  & 1.58(0.22)                    & 1.89(0.10)                     & 1.84(0.22)                    & 1.77(0.41)                     \\
mCCA-ICA                                              & 2.68(0.19)                                        & 1.82(0.21)                  & 1.67(0.17)                    & 1.47(0.26)                     & 1.97(0.13)                    & 1.61(0.29)                     \\ \hline
SAGPOOL                                               & 1.84(0.33)                                        & 1.55(0.06)                  & 1.44(0.13)                    & 1.52(0.18)                     & 1.50(0.24)                    & 1.74(0.23)                     \\
DIFFPOOL                                              & 1.27(0.20)                                      & 1.34(0.14)                  & 1.16(0.30)                    & 1.27(0.41)                     & 1.25(0.07)                    & 1.43(0.15)                     \\
Brain-Cheby                                           & 1.51(0.67)                                      & 1.17(0.26)                  & 1.24(0.31)                    & 0.79(0.06)                     & 1.09(0.21)                    & 1.58(0.41)                     \\
BrainNetCNN                                           & 1.26(0.19)                                        & 1.43(0.24)                  & \textbf{0.91(0.11)}           & 1.33(0.23)                     & 1.14(0.13)                    & 1.29(0.19)                     \\ \hline
Ours w/o Contra.                                      & \textbf{1.02(0.11)}                     & \textbf{0.89(0.13)}         & 0.97(0.20)                    & \textbf{0.74(0.17)}            & \textbf{0.96(0.15)}           & \textbf{1.15(0.11)}            \\
Ours                                                  & \textbf{0.83(0.24)}                     & \textbf{0.66(0.17)}         & \textbf{0.69(0.14)}           & \textbf{0.45(0.12)}            & \textbf{0.73(0.08)}           & \textbf{1.02(0.16)}            \\ \hline
\end{tabular}
\label{regression}
\end{table*}
\subsubsection{Experiment Setup}
In the regression tasks, we use the same baselines for comparisons.
The regression tasks include predicting MMSE scores on OASIS data, Flanker scores, Card-Sort scores, and $3$ ASR scores (i.e., Aggressive, Intrusive and Rule-Break scores) on HCP data. 
Particularly, MMSE (Mini-Mental State Exam) test \cite{tombaugh1992mini}, Flanker test \cite{eriksen1974effects} and Wisconsin Card-Sort test \cite{pangman2000examination,monchi2001wisconsin,berg1948simple} are $3$ neuropsychological tests designed to measure the status and risks of human neurodegenerative disease and mental illness.
The ASR (Achenbach Adult Self-Report) is a life function which is used to measure the emotion and social support of adults. 
The structure of proposed model remains unchanged. 
The loss weights are set as $\eta_{1}=0.5$ and $\eta_{2}=1$. 
The regression results are reported in average Mean Absolute Errors (MAE) with its \emph{std} under 5-fold cross validations.  

\subsubsection{Results}
The regression results are presented in Table \ref{regression}.  
It shows that our model achieves the best MAE values comparing to all baseline methods. 
Similar to the classification tasks, the deep graph neural networks are superior than traditional graph embedding methods (i.e., t-BNE and mCCA-ICA).
Comparing our method with and without the supervision of the contrastive loss, we can hold the conclusion that the contrastive learning can further boost the model performance. 
\subsection{Ablation Studies}
\label{abalation-studies}
In this section, we analyze the effect of $2$ hyperparameters on our model performance. 
The first parameter is the window size ($d$) which we used to clamp the fMRI bold signals when generating the contrastive functional brain networks. 
Particularly, we set the window size as  $[0, 5, 10, 20, 30, 40, 50]$, respectively and generate different contrastive samples as the input of our proposed model. 
Fig. \ref{para_win} shows the performance of our framework under different window sizes. 
It indicates that the best window size is around $d=10$. 
When the window size decreases to $0$, the model performance declines since the data is only duplicated without any substantial new samples.
It is interesting that the performance when $d=0$ is even worse than that obtained without contrastive learning but with contrastive samples generated with $d=10$ (see Ours w/o Contra. in Table \ref{classification} and \ref{regression}).
The reason is that data augmentation is introduced in the latter case, however, no augmented data is involved in the first case.  

We also analyze the effect of loss weights $\mu_{1}$ and $\mu_{2}$ on our model performance.
Fig. \ref{para_class_loss} presents the loss weight analysis on the $3$ classification tasks and the best results are achieved when $\mu_{1}=1$ and $\mu_{2}=0.1$. 
\subsection{Interpretation with Brain Saliency Map}
We utilize the Class Activation Mapping (CAM) approach \cite{zhang2020deep,arslan2018graph,pope2019explainability} to generate the brain network saliency map, which indicates the top brain regions associated with each prediction task.   
Figs \ref{SaliencyMaps_classification} and \ref{SalienceMaps_regression} illustrate Brain Saliency Maps for classification and regression tasks, respectively. 
For example, in the classification task (AD vs. NC), the saliency map for AD highlights multiple regions (such as Planum Polare, Frontal Operculum cortex, Supracalcarine Cortex, etc.) which are conventionally conceived as the biomarkers of AD in medical imaging analysis\cite{rasero2017multivariate,kutova2018asymmetric,hiscox2020mechanical,hafkemeijer2015resting}. 
In the meantime, the saliency map for NC highlights many regions in Cerebellum and Frontal lobe. These regions control cognitive thinking, motor control, and social mentalizing as well as emotional self-experiences\cite{stoodley2012functional,van2020posterior,sawyer2017diagnosing}, in which AD patients typically show problems. 
The details for all highlighted brain regions for each task are summarized in the Supplementary Material. 
These regions highlighted in the salience map can help us locating brain regions associated with any phenotype, which deserve further clinical investigations.


%% file: 06_conclusion.tex
\section{Conclusion}
We propose a novel contrastive learning framework with an interpretable hierarchical signed graph representation learning model for brain functional network mining. 
Additionally, a new data augmentation strategy is designed to generate the contrastive samples for brain functional network data. 
Our new framework is capable of generating more accurate representations for brain functional networks in compared with other state-of-the-art methods and these network representations can be used in various prediction tasks (e.g., classification and regression). Moreover, 
Brain saliency maps may assist with phenotypic biomarker identification and provide interpretable explanation on framework outcomes.

\section{Acknowledgement}
This study is partially supported by The National Institutes of Health (R01AG071243, R01MH125928 and U01AG068057) and National Science Foundation (IIS 2045848 and IIS 1837956).

Data were provided [in part] by the Human Connectome Project, MGH-USC Consortium (Principal Investigators: Bruce R. Rosen, Arthur W. Toga and Van Wedeen; U01MH093765) funded by the NIH Blueprint Initiative for Neuroscience Research grant; the National Institutes of Health grant P41EB015896, and the Instrumentation Grants S10RR023043, 1S10RR023401, 1S10RR019307.


\begin{table*}[h]
\caption{The ROI names of the highlighted brain regions in the saliency map in classification tasks}
\label{roi_name_class}
\resizebox{\textwidth}{45mm}{
\begin{tabular}{l|l|l|l|l|l|l}
\hline
\multicolumn{1}{c|}{\textbf{OASIS NC}}                                                         & \multicolumn{1}{c|}{\textbf{OASIS AD}}                                                          & \multicolumn{1}{c|}{\textbf{HCP Male}}                                 & \multicolumn{1}{c|}{\textbf{HCP Female}}                                  & \multicolumn{1}{c|}{\textbf{HCP Not Twins}}                           & \multicolumn{1}{c|}{\textbf{HCP Monozygotic}}                      & \multicolumn{1}{c}{\textbf{HCP Dizygotic}}                                \\ \hline
\begin{tabular}[c]{@{}l@{}}Paracingulate \\ Gyrus Right\end{tabular}                           & Planum Polare Left                                                                              & ctx-lh-precuneus                                                       & ctx-rh-superiorfrontal                                                    & \begin{tabular}[c]{@{}l@{}}ctx-lh-\\ lateraloccipital\end{tabular}    & \begin{tabular}[c]{@{}l@{}}ctx-lh-\\ isthmuscingulate\end{tabular} & ctx-lh-postcentral                                                        \\ \hline
\begin{tabular}[c]{@{}l@{}}Paracingulate \\ Gyrus Right\end{tabular}                           & \begin{tabular}[c]{@{}l@{}}Frontal Operculum \\ Cortex Left\end{tabular}                        & \begin{tabular}[c]{@{}l@{}}ctx-rh-\\ superiorparietal\end{tabular}     & Right-Accumbens-area                                                      & ctx-rh-bankssts                                                       & ctx-rh-pericalcarine                                               & \begin{tabular}[c]{@{}l@{}}ctx-lh-\\ caudalanteriorcingulate\end{tabular} \\ \hline
Frontal Pole Right                                                                             & \begin{tabular}[c]{@{}l@{}}Supracalcarine \\ Cortex Left\end{tabular}                           & Right-Hippocampus                                                      & \begin{tabular}[c]{@{}l@{}}ctx-rh-\\ caudalmiddlefrontal\end{tabular}     & ctx-rh-parsorbitalis                                                  & ctx-rh-parsorbitalis                                               & ctx-rh-parsorbitalis                                                      \\ \hline
Cerebellum 6 Right                                                                              & \begin{tabular}[c]{@{}l@{}}Superior Temporal \\ Gyrus, anterior \\ division Right\end{tabular}  & \begin{tabular}[c]{@{}l@{}}ctx-rh-\\ parahippocampal\end{tabular}      & ctx-lh-parsorbitalis                                                      & ctx-lh-precentral                                                     & ctx-rh-frontalpole                                                 & Right-Putamen                                                             \\ \hline
\begin{tabular}[c]{@{}l@{}}Paracingulate \\ Gyrus Left\end{tabular}                            & \begin{tabular}[c]{@{}l@{}}Supramarginal \\ Gyrus, anterior \\ division Right\end{tabular}      & Right-Amygdala                                                         & Right-Amygdala                                                            & \begin{tabular}[c]{@{}l@{}}ctx-lh-\\ parahippocampal\end{tabular}     & ctx-lh-fusiform                                                    & ctx-rh-precentral                                                         \\ \hline
Left-Putamen                                                                                   & Left-Caudate                                                                                    & ctx-lh-pericalcarine                                                   & ctx-rh-paracentral                                                        & ctx-lh-entorhinal                                                     & ctx-lh-entorhinal                                                  & \begin{tabular}[c]{@{}l@{}}ctx-rh-\\ caudalmiddlefrontal\end{tabular}     \\ \hline
Cerebellum 8 Left                                                                               & \begin{tabular}[c]{@{}l@{}}Middle Temporal \\ Gyrus, posterior \\ division Left\end{tabular}    & \begin{tabular}[c]{@{}l@{}}ctx-lh-\\ transversetemporal\end{tabular}   & ctx-lh-precentral                                                         & ctx-lh-superiorfrontal                                                & \begin{tabular}[c]{@{}l@{}}ctx-lh-\\ superiorfrontal\end{tabular}  & ctx-lh-precuneus                                                          \\ \hline
Cerebellum 7b Right                                                                             & \begin{tabular}[c]{@{}l@{}}Superior Temporal \\ Gyrus, posterior \\ division Left\end{tabular}  & \begin{tabular}[c]{@{}l@{}}ctx-rh-\\ transversetemporal\end{tabular}   & \begin{tabular}[c]{@{}l@{}}ctx-lh-\\ isthmuscingulate\end{tabular}        & Right-Pallidum                                                        & \begin{tabular}[c]{@{}l@{}}ctx-lh-\\ temporalpole\end{tabular}     & ctx-lh-temporalpole                                                       \\ \hline
Heschl's Gyrus Left                                                                            & Heschl's Gyrus Left                                                                             & \begin{tabular}[c]{@{}l@{}}ctx-rh-\\ lateralorbitofrontal\end{tabular} & \begin{tabular}[c]{@{}l@{}}ctx-rh-\\ isthmuscingulate\end{tabular}        & \begin{tabular}[c]{@{}l@{}}ctx-lh-\\ superiortemporal\end{tabular}    & \begin{tabular}[c]{@{}l@{}}ctx-lh-\\ superiorparietal\end{tabular} & \begin{tabular}[c]{@{}l@{}}ctx-rh-\\ transversetemporal\end{tabular}      \\ \hline
Cuneal Cortex Right                                                                            & \begin{tabular}[c]{@{}l@{}}Intracalcarine \\ Cortex Left\end{tabular}                           & \begin{tabular}[c]{@{}l@{}}ctx-lh-\\ temporalpole\end{tabular}         & \begin{tabular}[c]{@{}l@{}}ctx-lh-\\ caudalanteriorcingulate\end{tabular} & \begin{tabular}[c]{@{}l@{}}ctx-rh-\\ caudalmiddlefrontal\end{tabular} & Left-Pallidum                                                      & \begin{tabular}[c]{@{}l@{}}ctx-rh-\\ transversetemporal\end{tabular}      \\ \hline
\begin{tabular}[c]{@{}l@{}}Lateral Occipital \\ Cortex,\\ superior division Right\end{tabular} & \begin{tabular}[c]{@{}l@{}}Middle Frontal \\ Gyrus Left\end{tabular}                            &                                                                        &                                                                           &                                                                       &                                                                    &                                                                           \\ \hline
Precuneous Cortex                                                                              & Planum Polare Right                                                                             &                                                                        &                                                                           &                                                                       &                                                                    &                                                                           \\ \hline
Cerebellum Crus2 Left                                                                           & \begin{tabular}[c]{@{}l@{}}Temporal Fusiform \\ Cortex, anterior \\ division Left\end{tabular}  &                                                                        &                                                                           &                                                                       &                                                                    &                                                                           \\ \hline
Brain-Stem                                                                                     & \begin{tabular}[c]{@{}l@{}}Middle Temporal \\ Gyrus, temporooccipital \\ part Left\end{tabular} &                                                                        &                                                                           &                                                                       &                                                                    &                                                                           \\ \hline
Cerebellum 8 Right                                                                              & \begin{tabular}[c]{@{}l@{}}Supracalcarine \\ Cortex Right\end{tabular}                          &                                                                        &                                                                           &                                                                       &                                                                    &                                                                           \\ \hline
\end{tabular}}
\end{table*}

\begin{table*}[]
\caption{The ROI names of the highlighted brain regions in the saliency map in regression tasks}
\label{roi_name_regress}
\setlength{\tabcolsep}{2mm}{
\begin{tabular}{l|l|l|l|l|l}
\hline
\multicolumn{1}{c|}{\textbf{OASIS MMSE}}                                                         & \multicolumn{1}{c|}{\textbf{Flanker}}                              & \multicolumn{1}{c|}{\textbf{Card-Sort}}                                    & \multicolumn{1}{c|}{\textbf{Aggressive}}                           & \multicolumn{1}{c|}{\textbf{Intrusive}}                                   & \multicolumn{1}{c}{\textbf{Rule-Break}}                            \\ \hline
Right-Caudate                                                                                    & Left-Accumbens-area                                                & Left-Accumbens-area                                                        & ctx-lh-bankssts                                                    & ctx-lh-bankssts                                                           & ctx-lh-precuneus                                                   \\ \hline
Temporal Pole Right                                                                              & ctx-rh-fusiform                                                    & Left-Putamen                                                               & \begin{tabular}[c]{@{}l@{}}ctx-lh-\\ middletemporal\end{tabular}   & ctx-lh-parsorbitalis                                                      & ctx-lh-lingual                                                     \\ \hline
\begin{tabular}[c]{@{}l@{}}Middle Temporal \\ Gyrus, posterior \\ division Right\end{tabular}    & ctx-lh-inferiortemporal                                            & \begin{tabular}[c]{@{}l@{}}ctx-lh-\\ caudalmiddlefrontal\end{tabular}      & \begin{tabular}[c]{@{}l@{}}ctx-lh-\\ inferiortemporal\end{tabular} & \begin{tabular}[c]{@{}l@{}}ctx-lh-\\ inferiortemporal\end{tabular}        & \begin{tabular}[c]{@{}l@{}}ctx-lh-\\ inferiortemporal\end{tabular} \\ \hline
Cerebellum Crus1 Right                                                                            & ctx-rh-insula                                                      & ctx-rh-frontalpole                                                         & \begin{tabular}[c]{@{}l@{}}ctx-lh-\\ lateraloccipital\end{tabular} & \begin{tabular}[c]{@{}l@{}}ctx-lh-\\ parahippocampal\end{tabular}         & Right-Caudate                                                      \\ \hline
\begin{tabular}[c]{@{}l@{}}Temporal Occipital \\ Fusiform Cortex Left\end{tabular}               & \begin{tabular}[c]{@{}l@{}}ctx-lh-\\ middletemporal\end{tabular}   & \begin{tabular}[c]{@{}l@{}}ctx-lh-\\ rostralanteriorcingulate\end{tabular} & ctx-lh-precentral                                                  & \begin{tabular}[c]{@{}l@{}}ctx-lh-\\ caudalanteriorcingulate\end{tabular} & \begin{tabular}[c]{@{}l@{}}ctx-rh-\\ lateraloccipital\end{tabular} \\ \hline
Planum Temporale Left                                                                            & ctx-lh-postcentral                                                 & \begin{tabular}[c]{@{}l@{}}ctx-rh-\\ caudalmiddlefrontal\end{tabular}      & ctx-rh-temporalpole                                                & \begin{tabular}[c]{@{}l@{}}ctx-rh-\\ caudalmiddlefrontal\end{tabular}     & \begin{tabular}[c]{@{}l@{}}ctx-rh-\\ temporalpole\end{tabular}     \\ \hline
\begin{tabular}[c]{@{}l@{}}Middle Temporal \\ Gyrus, temporooccipital \\ part Left\end{tabular}  & ctx-lh-frontalpole                                                 & \begin{tabular}[c]{@{}l@{}}ctx-rh-\\ middletemporal\end{tabular}           & ctx-rh-frontalpole                                                 & \begin{tabular}[c]{@{}l@{}}ctx-rh-\\ supramarginal\end{tabular}           & \begin{tabular}[c]{@{}l@{}}ctx-lh-\\ supramarginal\end{tabular}    \\ \hline
\begin{tabular}[c]{@{}l@{}}Temporal Occipital \\ Fusiform Cortex Right\end{tabular}              & ctx-lh-temporalpole                                                & ctx-lh-frontalpole                                                         & ctx-rh-parsorbitalis                                               & ctx-rh-paracentral                                                        & ctx-rh-insula                                                      \\ \hline
Planum Temporale Right                                                                           & \begin{tabular}[c]{@{}l@{}}ctx-rh-\\ superiortemporal\end{tabular} & ctx-rh-precentral                                                          & \begin{tabular}[c]{@{}l@{}}ctx-rh-\\ parstriangularis\end{tabular} & \begin{tabular}[c]{@{}l@{}}ctx-rh-\\ parstriangularis\end{tabular}        & \begin{tabular}[c]{@{}l@{}}ctx-rh-\\ parstriangularis\end{tabular} \\ \hline
Frontal Orbital Cortex Left                                                                      & ctx-rh-precentral                                                  & ctx-rh-precuneus                                                           & ctx-rh-entorhinal                                                  & ctx-lh-precentral                                                         & Right-Amygdala                                                     \\ \hline
\begin{tabular}[c]{@{}l@{}}Middle Temporal \\ Gyrus, posterior \\ division Left\end{tabular}     &                                                                    &                                                                            &                                                                    &                                                                           &                                                                    \\ \hline
Vermis 9                                                                                         &                                                                    &                                                                            &                                                                    &                                                                           &                                                                    \\ \hline
\begin{tabular}[c]{@{}l@{}}Middle Temporal \\ Gyrus, temporooccipital \\ part Right\end{tabular} &                                                                    &                                                                            &                                                                    &                                                                           &                                                                    \\ \hline
Left-Caudate                                                                                     &                                                                    &                                                                            &                                                                    &                                                                           &                                                                    \\ \hline
Temporal Pole Left                                                                               &                                                                    &                                                                            &                                                                    &                                                                           &                                                                    \\ \hline
\end{tabular}}
\end{table*}